\documentclass[11pt,a4paper]{article}
\usepackage[hyperref]{eacl2021}
\usepackage{times}
\usepackage{latexsym}
\usepackage{comment}
\usepackage{graphicx}
\usepackage{amsmath}

\usepackage{microtype}

\aclfinalcopy 
\def\aclpaperid{35} 

\title{Misinformation detection in Luganda-English code-mixed social media text}

\author{Peter Nabende $^{1}$,
  David Kabiito $^{2}$, 
  Claire Babirye $^{2}$, 
  Hewitt Tusiime $^{2}$, \\
 \textbf{Joyce Nakatumba-Nabende} $^{3}$ \\
  $^{1}$ Makerere University, Department of Information Systems, \\ $^{2}$ Makerere University, Artificial Intelligence Lab, \\ $^{3}$ Makerere University, Department of Computer Science.}

\date{}

\begin{document}
\maketitle
\begin{abstract}
The increasing occurrence, forms, and negative effects of misinformation on 
social media platforms has necessitated more misinformation detection tools. Currently, work is being done addressing COVID-19 misinformation however, there are no misinformation detection tools for any of the 40 distinct indigenous Ugandan languages. This paper addresses this gap by presenting basic language resources and a misinformation detection data set based on code-mixed Luganda-English messages sourced from the Facebook and Twitter social media platforms. 
Several machine learning methods are applied on the misinformation detection data set to develop classification models for detecting whether a code-mixed Luganda-English message contains misinformation or not. A 10-fold cross validation evaluation of the classification methods in an experimental misinformation detection task shows that a Discriminative Multinomial Na\"ive Bayes (DMNB) method achieves the highest accuracy and F-measure of 78.19\% and 77.90\% respectively. Also, Support Vector Machine and Bagging ensemble classification models achieve comparable results. These results are promising since the machine learning models are based on n-gram features from only the misinformation detection data set.
\end{abstract}

\section{Introduction}

The Internet and the World Wide Web have provided a fertile ground to generate huge amounts of information thanks to various advantages and conveniences associated with them. Since the first blogging Web sites in 1999 where people started creating their own weblogs and sharing comments \cite{siles2012rise}, Web and mobile-based social media platforms have become popular, widely adopted, and have overtaken traditional media in facilitating various social interactions and aspects. As a result many social media platforms generate huge amounts of data and information. Various forms of social media content have proved useful in decision support applications in several areas such as business, health, and other services. However, due to the temptation to quickly post information and the lack of moderation \cite{jainetal2016}, more forms of social media content have turned out to be invalid, inaccurate, potentially harmful and in some cases intentionally harmful \cite{shuetal2017} to both individuals, groups, societies and other directly and indirectly affected entities. There are now several studies that have sought and continue to look for solutions to detect negative, wrong or undesired forms of information in social media posts. Most recent work has focused on addressing misinformation with regard to the Corona Virus disease 2019 (COVID-19) \cite{brennenetal2020,pennycooketal2020,serranoetal2020}. 

However, as is the case for several natural language processing (NLP) applications, there are hardly any low resourced languages that are involved in detecting misinformation, especially, in the current COVID-19 era. One major limitation for the lack of involvement of low resource languages are the few or no language resources such as labeled social media text, respective lexicons, and other basic NLP resources for developing misinformation detection models. In East Africa and Uganda in particular, the main effort towards detecting misinformation has focused on the analysis of speech data mined from community radios \cite{who-ug-2020,un-global-pulse-2019} to detect various forms of interesting information including misinformation. So far, it is mostly English and `Ugandan English' speech data that is analysed. This is the same case with another recent effort on sentiment classification of Twitter reviews about Ugandan Telecom companies \cite{kabiitoandnabende2020} where only English and `Ugandan English' were analysed. 

To the best of our knowledge there are no resources and misinformation detection models involving all of Uganda's over 40 distinct indigenous languages. In this paper, we present the first language resources including annotated social media text and pre-processing resources such as a polarity lexicon and other basic NLP resources for use in COVID-19 misinformation detection involving Uganda's most common indigenous language called Luganda. Luganda is also still a very low-resourced language in NLP terms although it has a considerably larger presence on social media platforms such as Twitter\footnote{\url{https://twitter.com}} and Facebook \footnote{\url{https://facebook.com}} compared to all other indigenous Ugandan languages. We utilize the Luganda language resources to evaluate various text classification methods for misinformation detection.

\section{Related Work}
Since the onset of the  COVID-19 pandemic, several studies have been carried out around the creation of COVID-19 misinformation datasets.  Fourati et al. (\citeyear{fourati2020tunizi})  create a Tunisian Arabizi sentiment analysis dataset. Elhadad et al. (\citeyear{elhadad2020covid}) provide an annotated bilingual Arabic to English COVID-19 dataset based on Twitter data. This dataset is useful for detecting misinformation around COVID-19 and is also similar to the work by Al-Zaman et al. (\citeyear{al2020covid}) where they develop a COVID-19 social media fake news dataset for India. Memon and Carley (\citeyear{memon2020characterizing}) provide a COVID-19 Twitter dataset in English that is classified into informed, misinformed and irrelevant groups. Cui and Lee (\citeyear{cui2020coaid}) creates a COVID-19 health care misinformation dataset that includes fake news on different websites and social media sites. 

In order to detect COVID-19 misinformation on social media data,  several techniques have been employed including the use of crowd sourcing to check for misinformation \cite{kim2020leveraging}, detection using
the elaborate likelihood models of persuasion from social media posts \cite{janze2017automatic} or building machine learning models for COVID-19 misinformation detection \cite{monti2019fake}. Patwa et al. (\citeyear{patwa2020fighting}) curate an annotated dataset of social media posts and articles of both real and fake news on COVID-19 and they also build four machine learning models (Decision Tree, Logistic Regression, Gradient Boost, and Support Vector Machine) with SVM giving the best model performance.  Similarly, Cui and Lee (\citeyear{cui2020coaid}) presents a machine learning model trained on a COVID-19 misinformation dataset which is in English.

The studies mentioned above underscore the potential of developing resources and solutions for identification of misinformation from social media. So far, there is hardly any work involving East African languages in detecting misinformation from social media. As efforts towards covering this gap, this study develops a preliminary misinformation detection dataset involving Luganda (a very low resourced East African language), and explores the application of several machine learning methods on the dataset.

\section{The Luganda language}
Luganda is a Bantu language and is the most widely spoken indigenous language in Uganda. It is  primarily spoken in the south eastern Buganda region  along the shores of Lake Victoria  and up north towards the shores of Lake Kyoga ~\cite{nakayiza2013sociolinguistics,olutola2019model}. According to a 2014 language census \cite{lewis2013ethnologue}, Luganda has over seven million first and second language speakers. Typologically, it is a highly-agglutinative, tonal language with subject-verb-object word order, and nominative-accusative morphosyntactic alignment \cite{olutola2019model}.

Although Luganda is the most widely spoken indigenous language in Uganda, it is considered a low resourced language because of the limited availability of Luganda NLP resources \cite{nandutu2020luganda}. The increased access of Internet subscriptions and internet-enabled phones in several parts of Uganda has led to an increase in the number of the World Wide Web users \cite{ucc2020}. A growing number of these users utilize social media platforms and usually post messages in Luganda and code-mixed Luganda-English \cite{ssentumbwe2020low}. The ever growing Luganda content on Web-based social media sites such as Twitter and Facebook opens opportunities for sourcing very much needed Luganda text for various text related NLP applications \cite{nannyongaetal2020}.

\section{Luganda-English code-mixed social media data}

Initially, the plan was to collect only Luganda text from social media Web sites; however, a manual analysis of a reasonable number of messages from the Web sources showed a lot of code switching between Luganda and English. This observation can be explained by the use of English as the main official language and the main mode of communication throughout the entire Ugandan Education system and Government. The speakers tend to code switch between their indigenous languages and English in informal communications such as messages posted on social media websites. The dataset used in this paper also included code switched Luganda-English text and this also enabled us to significantly increase the size of the dataset. Moreover, it is now established that code switching is very common and acceptable in multilingual communities \cite{begum2016functions}. The misinformation data used in this study are obtained from two commonly used social media sources, that is, Twitter and Facebook. During the data collection exercise, toxic or offensive statements that were not related to COVID-19 were dropped. However, toxic/offensive statements that discussed COVID-19 were not dropped and were annotated accordingly depending on whether they were true or false. As an example there were posts that dismissed the existence of COVID-19 in Uganda and called it a money making venture for government officials. Such posts are important because they contain misinformation about the existence of COVID-19; therefore, we included such posts and labeled them as having misinformation.
 
\subsection{Twitter data}

Using the Twitter API, we entered key words associated with Luganda broadcasting channels, political figures (for example \textit{cpmayiga} (the Prime Minister of the Buganda kingdom), general public messages (for example those with a hashtag \emph{StaySafeUG}) to retrieve Luganda and Luganda-English code-mixed tweets. We collected 12,049 raw tweets that had been posted from March 2020 to July 2020. The raw collection was pre-processed by converting all tweets to lowercase and removing the following: all mentions (\@twitter\_handle),  irrelevant abbreviations (such as rt for retweet), removing unnecessary symbols (such as the ampersand symbol (\&)) and  all non-ASCII characters. The pre-processing at this stage resulted into a total of 7,136 unlabeled tweets.

\subsection{Facebook data}
Using the  Facebook API, we extracted 430,075 posts between March 2020 and July 2020 from Facebook pages that belong to TV and radio broadcasting companies that mainly communicate using Luganda; these include: Bukedde TV, CBS, Spark TV, and Radio Simba. The Ugandan Ministry of Health had the main mandate to provide reliable information and guidance concerning the COVID-19 virus. So we also extracted Facebook posts from Uganda's Ministry of Health Facebook page (which is mainly in English) for the purpose of verifying the correctness of some information in the posts obtained from the Facebook pages of the luganda-based broadcasting companies.

\subsection{Topic modeling}

To gain more insight into the Luganda and Luganda-English code switch terms used in the Facebook and Twitter datasets, we used Latent Dirichlet Allocation (LDA) \cite{bleietal2003} to model topics from the unlabeled datasets. LDA is a well-established topic modeling algorithm that serves to identify topics that may be useful for tasks involving text classification. The dataset was preprocessed to remove stop words, punctuation, URLs, and any email addresses. One main development that arose at this stage was the creation of a list of Luganda stop words; this list was previously non-existent. This research has now developed a list of Luganda stop words that were used for preprocessing of the dataset. The datasets were further converted to lowercase before creating bi-grams and tri-grams that were used to train the LDA unsupervised model. Using the Gensim library, we trained the LDA model with the following set of hyper-parameters;  num\_topics:10, random\_state:100, chunksize:100, passes:10, alpha:'auto',  per\_word\_topics:True. 

We run the LDA models on the Twitter and Facebook data separately. Due to the high topic variation on the Facebook data, we got a perplexity score of 11.08 while that for twitter data was 9.01. These scores were obtained for 10 topics. When the Facebook and Twitter data was combined and the LDA topic modelling is done, a perplexity score of 13.66 was obtained with 10 topics. The increase in perplexity score is explained by the increase in topic variation introduced by combining the two data sources. With the increased topic variation the model finds it more complex to place the words under each topic.The topics were centered on the number of reported COVID-19 cases, government response to COVID-19, discussions on COVID-19 recommendations from prominent people like the Kabaka ``King'' of Buganda and Katikiro ``Prime Minister'', the origin of COVID-19, Presidential address on COVID-19, how the police was enforcing curfew, the pinch of the lockdown. Due to the high topic variation on the Facebook data, we got a perplexity score of 11.08 while the Twitter data gave a perplexity score of 9.01.

\begin{figure}[h]
  \centering
  \includegraphics[width=\linewidth]{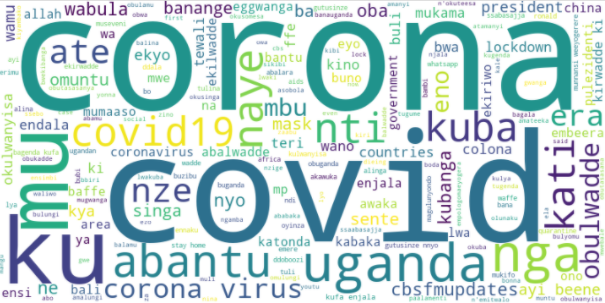}
  \caption{Facebook word cloud.}
  \label{facebook_wordcloud}
\end{figure}
\begin{figure}[h]
  \centering
  \includegraphics[width=\linewidth]{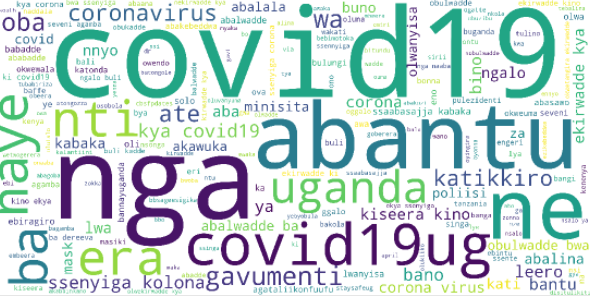}
  \caption{Twitter word cloud.}
  \label{twitter_wordcloud}
\end{figure}

Word clouds were determined for both the Twitter and Facebook data to provide a visual representation of the text from each of the platforms. A set of synonyms used to refer to COVID-19 on Facebook and Twitter was also obtained. Figure \ref{facebook_wordcloud} and \ref{twitter_wordcloud} show the word clouds associated with Facebook and Twitter datasets respectively. As shown in the word clouds, COVID-19 and corona are the most common words. It is important to note that COVID-19 in the datasets is synonymous with corona, virus, obulwadde, ekilwadde, coronavirus, covid19, kolona, covid, and ssenyiga.

\subsection{Labeling for the code-mixed COVID-19 misinformation dataset}

Annotation guidelines\footnote{Annotation guideline: \url{https://github.com/AI-Lab-Makerere/Luganda_misinformation/blob/master/data/annotation_guidelines.md}} were used to extract COVID-19 related tweets and posts written involving Luganda. The annotation was carried out by two annotators who were required to label each tweet or post as either ``misinformation" or ``no-misinformation". The annotators were given the same dataset to annotate during the training process. In case there was a tie, the difference in annotation was discussed with the trainer and subsequently all similar tweets or posts are annotated according to what was agreed on in the discussion. After the training exercise the annotators were given different sets of data to annotate. Each post or tweet had to be in Luganda exclusively or contain a code-mix of Luganda and English. We filtered out all the posts and comments that were entirely in English. After labeling, we had a total of 1,045 COVID-19 posts and tweets with the two class labels\footnote{Dataset:\url{https://github.com/AI-Lab-Makerere/Luganda_misinformation/blob/master/data/covid_facebook_twitter_Luganda.json}}. 719 posts were in Luganda entirely while 316 posts were code switched between Luganda and English.
As shown in Table \ref{datastatistics}, 286 labeled entries were from Twitter while 759 were from Facebook. 613 posts were labelled as ``misinformation" while 432 posts were labeled as ``no-misinformation". 76\% of the Facebook posts were labeled as ``misinformation" whereas only 12\% of the tweets were labeled as ``misinformation". However, during the data annotation, we did not indicate a label to show if the comments and posts were in either in Luganda or code-mixed for English-Luganda.  Most of the ``misinformation" posts were about the nonexistence of COVID-19 in Uganda. The second most popular set of misinformation posts were indicating that Ugandans can not catch the COVID-19 virus.

\begin{table}[ht]
\begin{tabular}{l c c}
\hline
\textbf{Data}  & \textbf{Twitter} & \textbf{Facebook} \\
\hline
Raw Data & 12,049 & 430,075 \\
\hline
Prepossessed Data & 7,136 & 114,130  \\
\hline
Code-mixed & 286 & 759 \\
 annotated data &  &   \\
\hline
Data with & 36 & 577  \\
misinformation &  &   \\
\hline
Data without  & 250 & 182  \\
misinformation &  &   \\
\hline
\end{tabular}
\caption{Statistics about the dataset.}
\label{datastatistics}
\end{table}

\section{Classification methods}

Given the Luganda-English code-mixed dataset, we utilized the following categories of methods for learning misinformation classification models: Bayesian methods, linear and non-linear functions, lazy classifiers, decision tree based methods, and meta (or ensemble) methods.

\subsection{Bayesian methods}

Na\"ive Bayes \cite{johnetal2013}, Bayesian Logistic Regression \cite{genkinetal2007}, and the Discriminative Multinomial Na\"ive Bayes classifier \cite{suetal2008}. The Na\"ive Bayes method usually serves as a baseline method in many text classification studies. It uses a strong independence assumption which may limit the representation of feature relationships in a dataset. More extended forms of Bayesian networks relax the strong assumption of Na\"ive Bayes but require more computational resources to be applied on cases with many features. Bayesian Logistic Regression extends classical logistic regression by allowing the use of prior information (or beliefs) to generate posterior distributions. The Discriminative Multinomial Na\"ive Bayes method uses a simple and efficient discriminative parameter learning method for multinomial Na\"ive Bayes network.

\subsection{Non-linear functions}

The Luganda-English code-mixed misinformation dataset is nonlinear. So we mainly use Support Vector Machines (SVMs) which are faster to implement and usually lead to competitively high performing classification models. The method of SVMs uses kernel functions (such as polynomial, sigmoid and radial basis) to transform a dataset from a nonlinear domain to a linearly separable domain. We explore various types of SVMs in the Liblinear \cite{fanetal2008} and LibSVM \cite{changandlin2001} libraries, in John Platt's Sequential Minimal Optimization algorithm \cite{platt1998}, and in the stochastic variant of the Pegasos (Primal Estimated sub-GrAdient SOlver for SVMs) method\cite{shalevetal2011}.

\subsection{Lazy classification methods}

Unlike previous methods that estimate a classification model based on training data, lazy classification methods keep the training data and only start to utilize it when they see a new instance for which they have to estimate a similarity with instances in the training data. The lazy classification methods explored here include: k-nearest neighbor \cite{ahaetal1991}, k-star \cite{clearyandtrigg1995}, and locally weighted learning \cite{atkesonetal1997}.

\subsection{Tree classification methods}

Tree classifiers are based on the notion of a decision tree where decision making considers different branches or paths to a final decisions in the leaves. By varying the characteristics of the tree branches and leaves, we get different types of trees. In this paper, we explore the following commonly used methods: the C4.5 algorithm \cite{quinlan2014c4} and its variations (C4.5 Consolidated \cite{ibargurenetal2015} and C4.5 graft \cite{webb1999}, logistic model trees \cite{landwehretal2005}, and random forests \cite{breiman2001}.

\section{Misinformation detection experimental setup}

\subsection{Feature generation, cross-validation setup and evaluation metrics}

Due to the size of the misinformation detection data set (1,045 instances) is considerably small, we used stratified 10 fold cross validation to evaluate the application of the different types of machine learning methods. The performance of the classification models was measured using accuracy, area under the receiver operating characteristic curve (AUROC), and f-measure. The accuracy is the percentage of correctly classified instances out of the total number of instances in the test set. The AUROC is a measure of the discriminative ability of a classifier. The f-measure is the harmonic mean of precision and recall. 

\subsection{Word n-gram Features}

We used the TweetToSparseFeatureVector filter \cite{kiritchenkoetal2014} to calculate sparse features based on word n-grams (unigrams, bigrams, and trigrams). CMU's TweetNLP tokenizer was used to map input text into a sparse feature vector based on only the text in the misinformation dataset. Table~\ref{datafeatures} shows the total number of features for each n-gram setting.

\begin{table}[ht]
\begin{tabular}{l c}
\hline
n-gram type & Number of features\\
\hline
uni-gram & 7347\\
bi-gram & 23748\\
tri-gram & 41547\\
\hline
\end{tabular}
\caption{Word n-gram feature sizes.}
\label{datafeatures}
\end{table}
 
\section{Results and Discussion}

Table \ref{unigramresults} and Table \ref{bigram-misinfo-results} show the results for the uni-gram and bi-gram feature settings respectively in the experimental Luganda-English code mixed misinformation detection task. Only results for the best performing model per method are shown. 

The best Liblinear model used an L2-regularized logistic regression (dual) model to achieve significantly higher scores compared to using the other types of SVMs. The best LibSVM model used a regularization parameter $\gamma$ with the value 0.4, a sigmoid kernel (equation 1), and a degree of 5 for the kernel.

\begin{equation}
K(X,Y) = \text{tanh}(\gamma \cdot X^{T} Y + r)
\end{equation}  

The best Sequential Minimization Optimization algorithm uses a logistic multinomial regression model with a ridge estimator as a \emph{calibrator} and a polynomial kernel (equation 2).

\begin{equation}
K(X,Y) = (\gamma \cdot X^{T} Y + r) ^{d} , \gamma > 0
\end{equation} 

The best SPegasos model used a log-loss function which leads to better evaluation scores than the Hinge loss function. The setting for the best Random forest model included an infinite maximum depth, and 500 iterations. For the ensemble methods (boosting (AdaBoostM1) and bagging), we used the DMNBtext classifier as the base classifier since it had the highest values for the different classification performance metrics. We found the classification performance to decrease with an increase in the number of iterations.  

\begin{table*}[ht]
\begin{center}
\begin{tabular}{|l|c|c|c|}
\hline
Classifier & Accuracy & AUROC & F-Measure\\
\hline
Na\"ive Bayes & 66.99 & 80.70 & 66.10\\
\hline
Bayesian Logistic Regression & 75.77 & 75.30 & 75.60\\
\hline
Discriminative Multinomial Na\"ive Bayes & \textbf{78.19} & 82.30 & \textbf{77.90}\\
\hline
Best Liblinear model & 77.12 & \textbf{82.70} & 76.90\\
\hline
Best LibSVM model & 76.45 & 75.90 & 76.20\\
\hline
Best Sequential Minimization Optimization & 76.16 & 75.50 & 75.80\\
\hline
Best SPegasos & 77.22 & 82.60 & 76.90\\
\hline
Best Random Forest & 76.54 & 82.60 & 75.80\\
\hline
Logistic Model trees & 73.46 & 77.40 & 73.20\\
\hline
Pruned C4.5 & 72.49 & 73.10 & 72.20\\
\hline
Pruned C4.5 using CTC algorithm & 70.56 & 69.60 & 70.40\\
\hline
C4.5 & 72.20 & 73.10 & 71.90\\
\hline
AdaboostM1 with DMNBtext & 76.06 & 81.10 & 75.80\\
\hline
Bagging with DMNBtext & 77.90 & 82.60 & 77.60\\
\hline
\end{tabular}
\caption{Misinformation detection results based on uni-gram features.}
\label{unigramresults}
\end{center}
\end{table*}

The best performing method in both cases is the DMNBtext classifier. Decision tree methods took longer to train and yet they do not achieve any improvement over the better performing Bayesian and SVM methods. So, we did not consider decision tree methods for both bi-gram and tri-gram settings.

\begin{table*}[ht]
\begin{center}
\begin{tabular}{|l|c|c|c|}
\hline
Classifier & Accuracy & AUROC & F-Measure\\
\hline
Na\"ive Bayes & 67.47 & 80.90 & 66.70\\ 
\hline
Bayesian Logistic Regression & 76.93 & 76.30 & 76.60\\ 
\hline
Discriminative Multinomial Na\"ive Bayes & \textbf{77.99} & 81.90 & \textbf{77.70}\\ 
\hline
Best Liblinear model & 77.32 & 76.80 & 77.10\\ 
\hline
Best LibSVM model & 76.83 & 76.20 & 76.50\\ 
\hline
Best Sequential Minimization Optimization & 76.54 & 75.80 & 76.10\\ 
\hline
Best SPegasos & 77.60 & \textbf{82.40} & 77.20\\ 
\hline
AdaboostM1 with DMNBtext & 75.48 & 79.00 & 75.20\\ 
\hline
Bagging with DMNBtext & 77.41 & 82.30 & 77.10\\ 
\hline
\end{tabular}
\caption{Misinformation detection results based on bi-gram features.}
\label{bigram-misinfo-results}
\end{center}
\end{table*}

We see no significant differences between misinformation detection results based on the uni-gram (Table \ref{unigramresults}) and bi-gram (Table \ref{bigram-misinfo-results}) feature settings. The same turned out to be the case for tri-gram feature settings where there was slight decline in performance across all models.

\subsection{Conclusion}
This paper set out to present two main contributions towards addressing the gap of the lack of language resources and solutions for detecting misinformation involving indigenous Ugandan languages. First of all, the paper presented the first carefully labeled misinformation dataset involving Luganda, a very low-resourced indigenous Ugandan language. Secondly, the dataset was used to evaluate several machine learning methods in an experimental Luganda-English code-mixed misinformation detection task. Evaluation results showed the highest and promising classification performances from the Discriminative Multinomial Na\"ive Bayes, the Support Vector Machine models from the liblinear library, and the ensemble models from the bagging method that used the DMNB and SVM models as base classifiers. In future, we will continue to add more content and language resources into the Luganda misinformation dataset, and investigate the application of more machine learning methods.

\section*{Acknowledgments}
The research is funded by a DSA 2020 Research Award to Makerere University.

\bibliography{anthology,eacl2021}
\bibliographystyle{acl_natbib}

\end{document}